\documentclass{article}

\usepackage[utf8]{inputenc} 
\usepackage[T1]{fontenc}    
\usepackage{hyperref}       
\usepackage{url}            
\usepackage{booktabs}       
\usepackage{nicefrac}       
\usepackage{microtype}      
\usepackage{xcolor}         
\usepackage{amsfonts,mathtools,amssymb,amsthm}
\usepackage{cancel}
\usepackage[ruled]{algorithm2e}
\usepackage{algorithmic}
\usepackage{caption,subcaption}
\usepackage{array}
\usepackage[normalem]{ulem}
\usepackage{multirow}

\theoremstyle{plain}
\newtheorem{proposition}{Proposition}[section]
\newtheorem{definition}{Definition}
\newtheorem{theorem}{Theorem}

\usepackage{wrapfig}

\allowdisplaybreaks

\title{The Triad of Failure Modes and a Possible Way Out}

\author{Emanuele Sansone \\
Department of Computer Science, KU Leuven \\
Leuven, Belgium \\
}
\date{}

\begin{document}

\maketitle

\begin{abstract}
We present a novel objective function for cluster-based self-supervised learning (SSL) that is designed to circumvent \textit{the triad of failure modes}, namely representation collapse, cluster collapse, and the problem of invariance to permutations of cluster assignments. This objective consists of three key components: (i) A generative term that penalizes representation collapse, (ii) a term that promotes invariance to data augmentations, thereby addressing the issue of label permutations and (ii) a uniformity term that penalizes cluster collapse. Additionally, our proposed objective possesses two notable advantages. Firstly, it can be interpreted from a Bayesian perspective as a lower bound on the data log-likelihood. Secondly, it enables the training of a standard backbone architecture without the need for asymmetric elements like stop gradients, momentum encoders, or specialized clustering layers. Due to its simplicity and theoretical foundation, our proposed objective is well-suited for optimization. Experiments on both toy and real world data demonstrate its effectiveness.
\end{abstract}

\section{Background}
\begin{wrapfigure}[12]{l}{0.4\linewidth}
     \centering
     \includegraphics[width=0.4\linewidth]{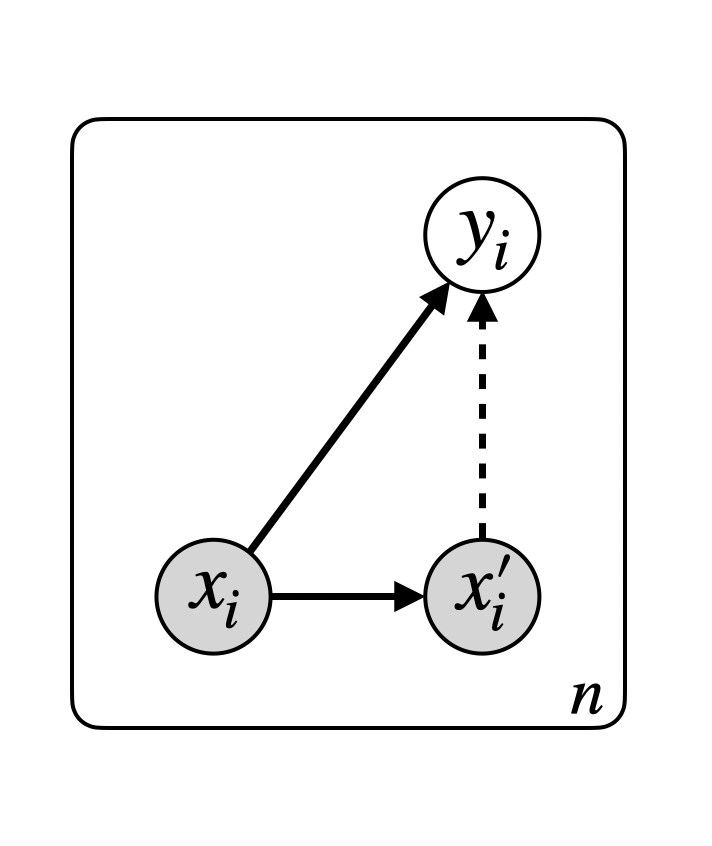}
     \caption{Probabilistic graphical model for cluster-based SSL. $i$ is used to index different training instances, i.e. $i=1,\dots,n$.}
     \label{fig:model}
\end{wrapfigure}
\textbf{Model}. Let us introduce the random quantities used in the model shown in Figure~\ref{fig:model}: (i) $x\in\Omega$, where $\Omega$ is a compact subset of $\mathbb{R}^d$, represents a data vector drawn independently from an unknown distribution $p(x)$ (for instance an image), (ii) $x'\in\Omega$ represents a transformed version of $x$ using a stochastic data augmentation strategy $\mathcal{T}(x'|x)$ (obtained by adding for instance noisy or cropping the original image), and (iii) $y\in\{1,\dots,c\}$ is the symbolic representation of an input data point defined over $c$ categories (namely the cluster label obtained by an output layer defined over the embedding representation). The corresponding probabilistic graphical model is given in Figure~\ref{fig:model}. The generative process (solid arrows) is defined using the following conditional densities, namely: $p(x'|x,\xi)=\mathcal{T}(x'|x)$ and $p(y|x)=\text{Softmax}(out(proj(enc(x))))$, where $enc:\Omega\rightarrow\mathbb{R}^h$ is an encoder used to compute the latent representation, $proj:\mathbb{R}^h\rightarrow\mathcal{S}^{h-1}$ is a projector head used to compute the embedding representation, and $out$ computes the cosine similarity between the embedding representation and the column vectors of a matrix of parameters $U\in\mathbb{R}^{h\times c}$ known as the cluster centers/prototypes~\cite{caron2020unsupervised}. The inference process (dashed arrow) is defined as $q(y|x)=\text{SK}(out(proj(enc(x'))))$, viz. a distribution over cluster/prototype assignments obtained through the Sinkhorn-Knopp algorithm (SK). Please refer to~\cite{caron2020unsupervised} for additional details.

\textbf{Objective}. The training objective is based on an evidence lower bound on the negative entropy, derived from the probabilistic graphical model of Figure~\ref{fig:model}(a), namely:
\begin{footnotesize}
\begin{align}
    \mathbb{E}_{p(x_{1:n})}\{\log p(x_{1:n};\Theta)\}
    & =-H_p(x_{1:n})+ \mathbb{E}_{p(x_{1:n})\mathcal{T}(x_{1:n}'|x_{1:n})}\left\{\log\sum_{y_{1:n}}p(y_{1:n}|x_{1:n};\Theta)\right\} \label{eq:general_cluster} \\
    & \geq -H_p(x_{1:n})+\underbrace{\sum_{i=1}^n\mathbb{E}_{p(x_i)\mathcal{T}(x_i'|x_i)}\left\{\mathbb{E}_{q(y_i|x_i')}\log p(y_i|x_i;\Theta) + H_q(y_i|x_i')\right\}}_\text{Discriminative term $\mathcal{L}_{DI}(\Theta)$}
\label{eq:discr_obj}
\end{align}
\end{footnotesize}
where $H_q(y|x')$ is the entropy computed over $q(y|x')$ and $\Theta$ includes all parameters of the encoder, projector head and the output layer of the discriminative model. Intuitively, the first addend in $\mathcal{L}_{DI}(\Theta)$ in Eq.~\ref{eq:discr_obj} forces the symbolic representations of the input data and its augmented version to be similar, whereas the second addend enforces uniformity on the cluster assignments, so as to avoid that all representations collapse to a single cluster. It is important to mention that the objective in Eq.~\ref{eq:discr_obj} is general enough to cover several proposed criteria in the literature of cluster-based self-supervised learning (cf.~\cite{sansone2023learning,sansone2022gedi}), such as DeepCluster~\cite{caron2018deep}, SwAV~\cite{caron2020unsupervised} and DINO~\cite{caron2021emerging}.

\section{Objective Function and The Triad of Failure Modes}
We devise a new lower bound for cluster-based SSL which avoids introducing asymmetries in the optimization procedure and in the discriminative backbone. We theoretically analyze the properties of the different loss terms involved in the GEDI instantiation with respect to important failure modes.

We are ready to state the following proposition (the proof can be found in Appendix A of the Supplementary Material):
\begin{proposition}
    \label{thm:lower_bound}
    Eq.~(\ref{eq:general_cluster}) can be lower bounded by the following quantity:
\begin{footnotesize}    \begin{align}
        -H_p(x_{1:n}) & \underbrace{- \sum_{i=1}^n \mathbb{E}_{p(x_i)\mathcal{T}(x_i'|x_i)}\left\{CE(p(y_i|x_i';\Theta),p(y_i|x_i;\Theta))\right\}}_\text{$\mathcal{L}_{INV}(\Theta)$} \underbrace{- \sum_{i=1}^nCE\left(p(y_i),q(y_i)\right)}_\text{$\mathcal{L}_{PRIOR}(\Theta)$}
        \label{eq:newbound}
    \end{align}
\end{footnotesize}
    with $q(y)=\frac{1}{n}\sum_{j=1}^np(y_j=y|x_j;\Theta)$ and $CE$ the cross-entropy loss.
    Additionally, the corresponding maximum value for the last two addends in Eq.~(\ref{eq:newbound}) is given by the following inequality:\footnote{Here, we assume that the predictive model $p(y|x;\Theta)$ has enough capacity to achieve the optimal solution.}
    \begin{footnotesize}
    \begin{align}
        \mathcal{L}_{INV}(\Theta)+\mathcal{L}_{PRIOR}(\Theta) \leq 
        &-H_p(y_{1:n}) \label{eq:maximum}
    \end{align}
    \end{footnotesize}
\end{proposition}
The above proposition has interesting implications. First of all, by maximizing the discriminative term $\mathcal{L}_{INV}(\Theta)$ with respect to $\Theta$, we enforce two properties, namely: (i) label invariance, as we ensure that the predictive distributions of the discriminative model for a sample and its augmented version match each other and (ii) confident predictions, as maximizing the cross-entropy forces also to decrease the entropy of these distributions.\footnote{Indeed, recall that $CE(p,q)=H_p + KL(p\|q)$. Therefore, maximizing $-CE(p,q)$ forces to have both $KL(p\|q)=0$ and $H_p=0$.} Secondly, by choosing a uniform prior, viz. $p(y_i)=\text{Uniform}(\{1,\dots,c\})$, and by maximizing $\mathcal{L}_{PRIOR}(\Theta)$ with respect to $\Theta$, we ensure to obtain a balanced cluster assignment, typical of approaches based on optimal transport objectives and corresponding surrogates~\cite{caron2018deep,caron2020unsupervised,cuturi2013sinkhorn}.
Finally, the proposed lower bound allows for an important key difference over existing cluster-based SSL, as we don't need to introduce asymmetries in the discriminative backbones. Indeed, we note that cluster-based SSL, specifically SwAV, assume $p(y|x;\Theta)=\text{Softmax}(U^Tg(x)/\tau)$ and $q(y|x')=\text{Sinkhorn}(\text{StopGrad}(U^Tg(x')/\tau))$, where $\text{Sinkhorn}$ and $\text{StopGrad}$ are two operators performing the Sinkhorn-Knopp algorithm and stopping the gradients, respectively. In contrast, we require that $q(y|x)=p(y|x;\Theta)=\text{Softmax}(f(enc(x))/\tau)$, where $f:\mathbb{R}^h\rightarrow\mathbb{R}^c$ is a simple discriminative network head.

%
%
Additionally, we lower bound the first addend in Eq.~\ref{eq:newbound} by exploiting the inequality $-H_p(x_{1:n})\geq-CE(p,p_\Theta)$, and obtain the overall objective, called GEDI (aka GEnerative DIscriminative objective):
\begin{footnotesize}
\begin{align}       \mathbb{E}_{p(x_{1:n})}\{\log p(x_{1:n};\Theta)\} \geq &\underbrace{\mathcal{L}_{GEN}(\Theta)}_\text{GEnerative term $-CE(p,p_\Theta)$} 
    + \underbrace{\mathcal{L}_{INV}(\Theta)+\mathcal{L}_{PRIOR}(\Theta)}_\text{DIscriminative terms}
    \label{eq:gedi_obj}
\end{align}
\end{footnotesize}
Importantly, we can reinterpret the discriminative model $p(y|x;\Theta)=\frac{p(y,x;\Theta)}{p(x;\Theta)}$ as an energy-based generative model $p_\Theta=p(x;\Theta)$, similarly to what is done in the context of supervised learning~\cite{grathwohl2020your,kim2022energy}, namely:
\begin{footnotesize}
\begin{align}
    p(y,x;\Theta)= \frac{e^{f_y(enc(x))/\tau}}{\Gamma(\Theta)} \qquad
    p_\Theta &\doteq p(x;\Theta)=\frac{\sum_{y=1}^c e^{f_y(enc(x))/\tau}}{\Gamma(\Theta)} = \frac{e^{\log\sum_{y=1}^c e^{f_y(enc(x))/\tau}}}{\Gamma(\Theta)}
    \label{eq:gedi_gen}
\end{align}
\end{footnotesize}
Training is performed by simply maximizing the lower bound in Eq~\ref{eq:gedi_obj}. We leave detailed discussion about the training and its computational requirements to Appendix B in the Supplementary Material. We are now ready to analyze the properties of the GEDI objective.

\textbf{The Triad of Failure Modes}. Here, we formalize three main failure modes for cluster-based SSL~\cite{wang2022importance}. Then, we study the GEDI loss landscape and show that these undesired trivial solutions are not admitted by our objective. This result holds without introducing asymmetries in the optimization procedure and/or network architecture.

Let's start by defining the most important failure modes, namely:
\begin{definition}[Failure Mode 1 - Representational Collapse] There exists a constant vector $k\in\mathbb{R}^h$ such that for all $x\in\mathbb{R}^d$, $enc(x)=k$.
\end{definition}
\begin{definition}[Failure Mode 2 - Cluster Collapse] There exists a cluster $j\in\{1,\dots,c\}$ such that for all $x\in\mathbb{R}^d$, $p(y=j|x;\Theta)=1$.
\end{definition}
\begin{definition}[Failure Mode 3 - Permutation Invariance to Cluster Assignments] For all possible permutations $\pi:\{1,\dots,c\}\rightarrow\{1,\dots,c\}$, a dataset $\mathcal{D}=\{(x_i,t_i,t_i')\}_{i=1}^n$, its permuted version $\mathcal{D}^\pi=\{(x_i,t_{\pi(i)},t_i')\}_{i=1}^n$ and a loss $\mathcal{L}(\Theta;\cdot)$, evaluated at one of the two datasets, we have that $\mathcal{L}(\Theta;\mathcal{D})=\mathcal{L}(\Theta;\mathcal{D}^\pi)$. For GEDI, $t_i\doteq f(enc(x_i))$  and $t_i'\doteq f(enc(x_i'))$.
\end{definition}
In other words, Definition 1 considers the case where the encoder maps (collapses) every input to the same output. Definition 2 considers the situation where the predictive model assigns all samples to the same cluster with high confidence. And Definition 3 considers the case where a hypothetical adversary swaps the predictions made by the model on different pair of inputs. Ideally, we would like to have an objective that does not admit these failure modes.

Now, we state the properties of the loss landscape of GEDI with the following theorem (we leave the proof to Section G in the Supplementary Material):
\begin{theorem}
\label{thm:admissible}
    Given definitions 1-3, the following statements tells for a particular loss, which modes are admitted as optimal solutions:
    \begin{itemize}
        \item[a.] $\mathcal{L}_{GEN}(\Theta)$ admits failure modes 2 and 3.
        \item[b.] $\mathcal{L}_{INV}(\Theta)$ admits failure modes 1 and 2.
        \item[c.] $\mathcal{L}_{PRIOR}(\Theta)$ admits failure modes 1 and 3.
    \end{itemize}
\end{theorem}
Importantly, Theorem~\ref{thm:admissible} tells us that $\mathcal{L}_{GEN}(\Theta)$ can be used to penalize representational collapse, $\mathcal{L}_{INV}(\Theta)$ can be used to break the problem of permutation invariance for the cluster assignments, while $\mathcal{L}_{PRIOR}(\Theta)$ can be used to penalize cluster collapse. Consequently, by maximizing the objective in Eq.~(\ref{eq:gedi_obj}), we are guaranteed to learn solutions which are non-trivial. A table summarizing all these properties is given below.
\begin{table}[h]
  \caption{Summary of loss landscape}
  \label{tab:loss}
  \centering
\resizebox{0.8\linewidth}{!}{\begin{tabular}{@{}lrrr@{}}
\toprule
\textbf{Does $\downarrow$ penalize $\rightarrow$?} & \textbf{Repr. collapse} & \textbf{Clus. collapse} & \textbf{Perm. Inv.} \\ 
\midrule
$\mathcal{L}_{GEN}(\Theta)$ & \textbf{Yes} & No & No \\
$\mathcal{L}_{INV}(\Theta)$ & No & No & \textbf{Yes} \\
$\mathcal{L}_{PRIOR}(\Theta)$ & No & \textbf{Yes} & No \\
Eq.~(\ref{eq:gedi_obj}) & \textbf{Yes} & \textbf{Yes} & \textbf{Yes} \\
\bottomrule
\end{tabular}}
\end{table}

\newpage
\section{Experiments}

\begin{table*}
  \caption{Clustering performance based on normalized mutual information (NMI) on test set (toy data, viz. moons and circles, and real data, viz. SVHN, CIFAR-10, CIFAR-100). Higher values indicate better clustering performance. Mean and standard deviations are computed from 5 different runs.}
  \label{tab:nmi_toy}
  \centering
\resizebox{\linewidth}{!}{\begin{tabular}{@{}lrrrrrr@{}}
\toprule
\textbf{Dataset} & \textbf{JEM}~\cite{grathwohl2020your} & \textbf{SwAV}~\cite{caron2020unsupervised} & \textbf{GEDI no unif} & \textbf{GEDI no inv} &  \textbf{GEDI no gen} & \textbf{GEDI} \\
\midrule
Moons & 0.00$\pm$0.00 &0.76$\pm$0.36 & 0.00$\pm$0.00 & 0.11$\pm$0.15 & \textbf{0.98}$\pm$\textbf{0.00} & 0.94$\pm$0.07 \\
Circles & 0.00$\pm$0.00 & 0.00$\pm$0.00 & 0.00$\pm$0.00 & 0.22$\pm$0.13 & 0.83$\pm$0.12 &
\textbf{1.00}$\pm$\textbf{0.01} \\
\midrule
SVHN & 0.00 & 0.21 & - & - & 0.21 & \textbf{0.25} \\
CIFAR10 & 0.00 & 0.43 & - & - & 0.43 & \textbf{0.45} \\
CIFAR100 & 0.00 & 0.65 & - & - & 0.86 & \textbf{0.87} \\
\bottomrule
\end{tabular}}
\end{table*}
We perform experiments to evaluate the discriminative performance of GEDI and its competitors, namely an energy-based model JEM~\cite{grathwohl2020your} and a self-supervised baseline based on SwAV~\cite{caron2020unsupervised}.
The whole analysis is divided into two main experimental settings, the first one based on two synthetic datasets, including moons and circles, the second one based on real-world data, including SVHN, CIFAR-10 and CIFAR-100. We use existing code both as a basis to build our solution and also to run the experiments for the different baselines. In particular, we use the code from~\cite{duvenaud2021no} for training energy-based models and the repository from~\cite{costa2022solo} for all self-supervised baselines. Implementation details as well as additional experiments on generation, OOD detection and linear probe evaluation are reported in the Supplementary Material (Appendices D-G).

\textbf{Moons and Circles}.In Table~\ref{tab:nmi_toy}, we observe that JEM fails to solve the clustering task for both datasets. This is quite natural, as JEM is a purely generative approach, mainly designed to perform implicit density estimation. SwAV can only solve the clustering task for the moons dataset, highlighting the fact that its objective function admits failure mode 3. Indeed, we observe in the circles dataset that half of the labels are permuted across the two manifolds (cf.  Figure~\ref{fig:labels_toy} in the Supplementary Material). In contrast, GEDI can recover the true clusters in both datasets, as it is guaranteed to avoid trivial solutions and learn more meaningful cluster assignments. We conduct an ablation study to understand the impact of the different loss terms in GEDI and empirically validate the theoretical results obtained in Section 4.3. We compare four different versions of GEDI, namely the full version (called simply GEDI), GEDI trained without $\mathcal{L}_{GEN}(\Theta)$ (called \textit{no gen}), GEDI trained without $\mathcal{L}_{INV}(\Theta)$ (called \textit{no inv}) and GEDI trained without $\mathcal{L}_{PRIOR}(\Theta)$ (called \textit{no unif}). From the results in Table~\ref{tab:nmi_toy}, we observe that: (i) GEDI \textit{no unif} is subject to cluster collapse on both datasets. This is expected as failure mode 2 is not penalized during training due to the omission of $\mathcal{L}_{PRIOR}(\Theta)$; (ii) GEDI \textit{no inv} is subject to the problem of permutation invariance to cluster assignments. Consequently, the obtained cluster labels are not informative and consistent with the underlying manifold structure of the data distribution. Again, this confirms the result of Theorem~\ref{thm:admissible}, as failure mode 3 could be avoided by the use of $\mathcal{L}_{INV}(\Theta)$; (iii) GEDI \textit{no gen} achieves superior performance over other SSL baselines. While in theory the objective function for this approach admits representational collapse, in practice we never observed such issue. It might be the case that the learning dynamics of gradient-based optimisation are enough to avoid the convergence to this trivial solution. However, further analysis is required in order to verify this statement; finally (iv) GEDI is guaranteed to avoid the most important failure modes and therefore solve the discriminative task. 

\textbf{SVHN, CIFAR-10, CIFAR-100}. From Table~\ref{tab:nmi_toy}, we observe that GEDI is able to outperform all other competitors by a large margin. Additionally, we note a difference gap in clustering performance with increasing number of classes (cf. CIFAR-100). This might be explained by the fact that the number of possible label permutations increases with the number of classes and that our loss is more robust to the permutation invariance problem as from Theorem~\ref{thm:admissible}. Finally,  GEDI \textit{no gen} is comparable and often superior to SwAV, despite being simpler (i.e. avoiding the use of asymmetries and the running of iterative clustering). Please refer to Appendices F and G for further details.

\bibliography{ref}

\begin{thebibliography}{10}

\bibitem{caron2018deep}
M.~Caron, P.~Bojanowski, A.~Joulin, and M.~Douze.
\newblock {Deep Clustering for Unsupervised Learning of Visual Features}.
\newblock In {\em ECCV}, 2018.

\bibitem{caron2020unsupervised}
M.~Caron, I.~Misra, J.~Mairal, P.~Goyal, P.~Bojanowski, and A.~Joulin.
\newblock {Unsupervised Learning of Visual Features by Contrasting Cluster Assignments}.
\newblock In {\em NeurIPS}, 2020.

\bibitem{caron2021emerging}
M.~Caron, H.~Touvron, I.~Misra, H.~J{\'e}gou, J.~Mairal, P.~Bojanowski, and A.~Joulin.
\newblock {Emerging Properties in Self-Supervised Vision Transformers}.
\newblock In {\em ICCV}, 2021.

\bibitem{cuturi2013sinkhorn}
M.~Cuturi.
\newblock {Sinkhorn Distances: Lightspeed Computation of Optimal Transport}.
\newblock In {\em NeurIPS}, 2013.

\bibitem{costa2022solo}
V.~G.~T. da~Costa, E.~Fini, M.~Nabi, N~Sebe, and E.~Ricci.
\newblock {Solo-learn: A Library of Self-supervised Methods for Visual Representation Learning}.
\newblock {\em JMLR}, 2022.

\bibitem{du2019implicit}
Y.~Du and I.~Mordatch.
\newblock {Implicit Generation and Generalization in Energy-Based Models}.
\newblock {\em arXiv}, 2019.

\bibitem{duvenaud2021no}
D.~Duvenaud, J.~Kelly, K.~Swersky, M.~Hashemi, M.~Norouzi, and W.~Grathwohl.
\newblock {No MCMC for Me: Amortized Samplers for Fast and Stable Training of Energy-Based Models}.
\newblock In {\em ICLR}, 2021.

\bibitem{grathwohl2020your}
W.~Grathwohl, K.-C. Wang, J.-H. Jacobsen, D.~Duvenaud, M.~Norouzi, and K.~Swersky.
\newblock {Your Classifier is Secretly an Energy Based Model and You Should Treat It Like One}.
\newblock In {\em ICLR}, 2020.

\bibitem{heusel2017gans}
M.~Heusel, H.~Ramsauer, T.~Unterthiner, B.~Nessler, and S.~Hochreiter.
\newblock {GANs Trained by a Two Time-Scale Update Rule Converge to a Local Nash Equilibrium}.
\newblock In {\em NeurIPS}, 2017.

\bibitem{kim2022energy}
B.~Kim and J.~C. Ye.
\newblock {Energy-Based Contrastive Learning of Visual Representations}.
\newblock In {\em NeurIPS}, 2022.

\bibitem{nalisnick2019deep}
E.~Nalisnick, A.~Matsukawa, Y.~W. Teh, D.~Gorur, and B.~Lakshminarayanan.
\newblock Do deep generative models know what they don't know?
\newblock In {\em ICLR}, 2019.

\bibitem{nijkamp2020anatomy}
E.~Nijkamp, M.~Hill, T.~Han, S.-C. Zhu, and Y.~N. Wu.
\newblock {On the Anatomy of MCMC-Based Maximum Likelihood Learning of Energy-Based Models}.
\newblock In {\em AAAI}, 2020.

\bibitem{nijkamp2019learning}
E.~Nijkamp, M.~Hill, S.~C. Zhu, and Y.~N. Wu.
\newblock Learning non-convergent non-persistent short-run mcmc toward energy-based model.
\newblock In {\em NeurIPS}, 2019.

\bibitem{sansone2022gedi}
E.~Sansone and R.~Manhaeve.
\newblock {GEDI: GEnerative and DIscriminative Training for Self-Supervised Learning}.
\newblock {\em arXiv}, 2022.

\bibitem{sansone2023learning}
E.~Sansone and R.~Manhaeve.
\newblock {Learning Symbolic Representations Through Joint GEnerative and DIscriminative Training}.
\newblock In {\em ICLR Workshop NeSy-GeMs}, 2023.

\bibitem{wang2022importance}
X.~Wang, H.~Fan, Y.~Tian, D.~Kihara, and X.~Chen.
\newblock {On the Importance of Asymmetry for Siamese Representation Learning}.
\newblock In {\em CVPR}, 2022.

\bibitem{xie2016theory}
J.~Xie, Y.~Lu, S.~C. Zhu, and Y.~N. Wu.
\newblock {A Theory of Generative Convnet}.
\newblock In {\em ICML}, 2016.

\end{thebibliography}
\bibliographystyle{plain}

\appendix
{\section{Proof of Proposition~\ref{thm:lower_bound}}\label{sec:proof_bound}}
\begin{proof}
We recall Eq.~(\ref{eq:general_cluster}) (we omit the dependence from $\Theta$ to avoid clutter), namely:
\begin{align}
    &\mathbb{E}_{p(x_{1:n})}\{\log p(x_{1:n})\}\nonumber\\
& =-H_p(x_{1:n})+ \mathbb{E}_{p(x_{1:n})\mathcal{T}(x_{1:n}'|x_{1:n})}\left\{\log\sum_{y_{1:n}}p(y_{1:n}|x_{1:n})\right\}\nonumber
\end{align}
and add the zero quantity $\log\sum_{y_{1:n}}p(y_{1:n})$ to the right-hand side of previous equation, thus obtaining the new equation
\begin{align}
    &\mathbb{E}_{p(x_{1:n})}\{\log p(x_{1:n})\}\nonumber\\
& =-H_p(x_{1:n})+ \mathbb{E}_{p(x_{1:n})\mathcal{T}(x_{1:n}'|x_{1:n})}\left\{\log\sum_{y_{1:n}}p(y_{1:n}|x_{1:n})\right\}\nonumber\\
&\qquad + \log\sum_{y_{1:n}}p(y_{1:n})
\label{eq:last}
\end{align}
We can lower bound the previous equation by exploiting the fact that $\sum_{z}p(z)\geq \sum_{z}p(z)q(z)$ for any given auxiliary discrete distribution $q$, viz.:
\begin{align}
&\text{Eq.~(\ref{eq:last})} \geq \nonumber\\
&\qquad -H_p(x_{1:n}) \nonumber\\
&\qquad +\mathbb{E}_{p(x_{1:n})\mathcal{T}(x_{1:n}'|x_{1:n})}\left\{\log\sum_{y_{1:n}}q(y_{1:n}|x_{1:n}')p(y_{1:n}|x_{1:n})\right\} \nonumber\\
&\qquad + \log\sum_{y_{1:n}}p(y_{1:n})q(y_{1:n})
\label{eq:last2}
\end{align}
Now, by applying Jensen's inequality to the last two addends in Eq.~(\ref{eq:last2}) and by defining $q(y_{1:n}|x_{1:n}')=p(y_{1:n}|x_{1:n}')$ and $q(y_{1:n})=\frac{1}{n}\sum_{j=1}^np(y_j|x_j)$, we obtain the following lower bound:
\begin{align}
    & \text{Eq.~(\ref{eq:last2})} \geq \nonumber \\
    &\qquad -H_p(x_{1:n}) \nonumber\\
    &\qquad +\mathbb{E}_{p(x_{1:n})\mathcal{T}(x_{1:n}'|x_{1:n})}\left\{\sum_{y_{1:n}}p(y_{1:n}|x_{1:n}')\log p(y_{1:n}|x_{1:n})\right\} \nonumber\\
    &\qquad + \sum_{y_{1:n}}p(y_{1:n})\log\left(\frac{1}{n}\sum_{j=1}^np(y_j|x_j)\right)
    \label{eq:last3}
\end{align}
Additionally, by factorizing the distributions according to the probabilistic graphical model in Fig.~\ref{fig:model}, namely $p(y_{1:n}|x_{1:n})=\prod_{i=1}^n p(y_i|x_i)$, $p(y_{1:n}|x_{1:n}')=\prod_{i=1}^n p(y_i|x_i')$ and $p(y_{1:n})=\prod_{i=1}^np(y_i)$, we achieve the following equality:
\begin{align}
    & \text{Eq.~(\ref{eq:last3})} = \nonumber\\
    &\qquad -H_p(x_{1:n}) \nonumber\\
    &\qquad +\sum_{i=1}^n\mathbb{E}_{p(x_i)\mathcal{T}(x_i'|x_i)}\left\{\sum_{y_i}p(y_i|x_i')\log p(y_i|x_i)\right\} \nonumber\\
    &\qquad + \sum_{i=1}^n\sum_{y_i}p(y_i)\log\left(\frac{1}{n}\sum_{j=1}^np(y_j=y_i|x_j)\right)
    \label{eq:last4}
\end{align}
And by rewriting the last two addends in Eq.~(\ref{eq:last4}) using the definition of cross-entropy, we obtain our final result.

Now, we can conclude the proof by looking at the maxima for $\mathcal{L}_{INV}$ and $\mathcal{L}_{PRIOR}$. Indeed, we observe that both terms compute a negative cross-entropy between two distributions. By leveraging the fact that $CE(p,q)=H_p + KL(p\|q)$ for arbitrary distributions $p,q$, we can easily see that the maximum of $\mathcal{L}_{INV}$ is attained when the term is $0$ (corresponding to minimal entropy and minimal KL), whereas the maximum of $\mathcal{L}_{PRIOR}$ is attained when the term is equal to $-H_p(y_i)$ (corresponding to minimal KL).
\end{proof}

{\section{Training Algorithm and Computational Requirements}}
\textbf{Learning a GEDI model.} We can train the GEDI model by jointly maximizing the objective in Eq.~(\ref{eq:gedi_obj}) with respect to the parameters $\Theta$ through gradient-based strategies. The overall gradient includes the summation of three terms, viz. $-\nabla_\Theta CE(p,p_\Theta)$, $\nabla_\Theta\mathcal{L}_{INV}(\Theta)$ and $\nabla_\Theta\mathcal{L}_{PRIOR}(\Theta)$. While the last two gradient terms can be computed easily by leveraging automatic differentiation, the first one must be computed by exploiting the following identities (obtained by simply substituting Eq.~(\ref{eq:gedi_gen}) into the definition of cross-entropy and expanding $\nabla_\Theta\Gamma(\Theta)$):
\begin{align}
    -\nabla_\Theta CE(p,p_\Theta) &= \sum_{i=1}^n\mathbb{E}_{p(x_i)}\left\{\nabla_\Theta\log\sum_{y=1}^c e^{f_y(enc(x_i))/\tau}\right\} \nonumber\\
    &\quad-n\nabla_\Theta\log\Gamma(\Theta) \nonumber\\
    &=\sum_{i=1}^n\mathbb{E}_{p(x_i)}\left\{\nabla_\Theta\log\sum_{y=1}^c e^{f_y(enc(x_i))/\tau}\right\}  \nonumber\\
    &\quad n\mathbb{E}_{p_\Theta(x)}\left\{\nabla_\Theta\log\sum_{y=1}^c e^{f_y(enc(x))/\tau}\right\}
    \label{eq:energy_gedi}
\end{align}
Importantly, the first and the second expectations in Eq.~(\ref{eq:energy_gedi}) are estimated using the training and the generated data, respectively. To generate data from $p_\Theta$, we use a sampler based on Stochastic Gradient Langevin Dynamics (SGLD), thus following recent best practices to train energy-based models~\cite{xie2016theory,nijkamp2019learning,du2019implicit,nijkamp2020anatomy}.
\begin{algorithm}[t]
 \caption{GEDI Training.}
 \label{alg:gedi}
    \textbf{Input:} $x_{1:n}$, $x_{1:n}'$, $\text{Iters}$, SGLD and Adam optimizer hyperparameters\;
    \textbf{Output:} Trained model $\Theta$\;
    \textbf{For} $\text{iter}=1,\dots,\text{Iters}$\;
    \qquad Generate samples from $p_\Theta$ using SGLD\;
    \qquad Estimate $\Delta_1\Theta=\nabla_\Theta \mathcal{L}_{GEN}(\Theta)$ using Eq.~(\ref{eq:energy_gedi})\;
    \qquad Compute $\Delta_2\Theta=\nabla_\Theta\mathcal{L}_{INV}(\Theta)$\;
    \qquad Compute $\Delta_3\Theta=\nabla_\Theta\mathcal{L}_{PRIOR}(\Theta)$ \;
    \qquad $\Delta\Theta\leftarrow \sum_{i=1}^3\Delta_i\Theta$\;
    \qquad $\Theta\leftarrow\text{Adam}$ maximizing using $\Delta\Theta$\;
    \textbf{Return} $\Theta$\;
\end{algorithm}
The whole learning procedure is summarized in Algorithm~\ref{alg:gedi}.

\textbf{Computational requirements.} When comparing our GEDI instantiation with traditional SSL training, more specifically to SwAV, we observe two main differences in terms of computation. Firstly, our learning algorithm does not require to run the Sinkhorn-Knopp algorithm, thus saving computation. Secondly, our GEDI instantiation requires additional forward and backward passes to draw samples from the energy-based model $p_\Theta$. However, the number of additional passes through the discriminative model can be limited by the number of SGLD iterations.

{\section{Proof of Theorem~\ref{thm:admissible}}\label{sec:admissible}}
\begin{proof}
The overall strategy to prove the statements relies on the evaluation of the loss terms over the three failure modes and on checking whether these attain their corresponding maxima.

Let's start by proving statement a and recalling that $\mathcal{L}_{GEN}(\Theta,\mathcal{D})=-CE(p,p_\Theta)$. Firstly, we test for failure mode 1 (i.e. representational collapse). We observe that for all $x\in\mathbb{R}^d$ 
$$p_\Theta(x)=\frac{\sum_{y=1}^ce^{f_y(k)/\tau}}{\Gamma(\Theta)}$$
thus $p_\Theta(x)$ assigns constant mass everywhere. Clearly, $p_\Theta$ is different from $p$. Therefore, $-CE(p,p_\Theta)<-CE(p,p)$ and failure mode 1 is not admissible. Secondly, we test for failure mode 2 (i.e. cluster collapse). We can equivalently rewrite the definition of cluster collapse by stating that there exists $j\in\{1,\dots,c\}$, such that for all $x\in\mathbb{R}^d$ and for all $y\neq j$, $f_j(enc(x))-f_y(enc(x))\rightarrow\infty$. Additionally, we observe that
\begin{align}
    p_\Theta(x) &\underbrace{=}_\text{$\xi_x\doteq enc(x)$}\frac{\sum_{y=1}^ce^{f_y(\xi_x)/\tau}}{\int \sum_{y=1}^ce^{f_y(\xi_x)/\tau}dx} \nonumber\\
    &=\frac{e^{f_j(\xi_x)/\tau}\left[1+\sum_{y\neq j}\cancelto{0}{e^{(f_y(\xi_x)-f_j(\xi_x))/\tau}}\right]}{\int e^{f_j(\xi_x)/\tau}\left[1+\sum_{y\neq j}\cancelto{0}{e^{(f_y(\xi_x)-f_j(\xi_x))/\tau}}\right] dx} \nonumber\\
    &=\frac{e^{f_j(\xi_x)/\tau}}{\int e^{f_j(\xi_x)/\tau}dx} 
    \label{eq:temp_ebm}
\end{align}
where we have used the failure mode condition to obtain the last equality. Now, note that Eq.~(\ref{eq:temp_ebm}) defines a standard energy-based model. Consequently, given enough capacity for the predictive model, it is trivial to verify that there exists $\Theta$ such that the condition about failure mode is met and $p_\Theta$ is equal to $p$. Cluster collapse is therefore an admissible solution.  Thirdly, we test for permutation invariance for the cluster assignments. Indeed, we have that
\begin{align}
    \mathcal{L}_{GEN}(\Theta,\mathcal{D}) & =\sum_{i=1}^n\mathbb{E}_{p(x_i)}\left\{\log p_\Theta(x_i)\right\} \nonumber\\
    &=\sum_{i=1}^n\mathbb{E}_{p(x_i)}\left\{\log \frac{\sum_{y=1}^ce^{t_i(y)/\tau}}{\int \sum_{y=1}^ce^{t_i(y)/\tau}dx}\right\} \nonumber \\
    &=\sum_{i=1}^n\mathbb{E}_{p(x_{1:n})}\left\{\log \frac{\sum_{y=1}^ce^{t_i(y)/\tau}}{\int \sum_{y=1}^ce^{t_i(y)/\tau}dx}\right\}
    \label{eq:lgen}
\end{align}
where $t_i(y)=f_y(enc(x_i))$.
Similarly, we have that
\begin{align}
    &\mathcal{L}_{GEN}(\Theta,\mathcal{D}^\pi) \nonumber\\
    &\qquad\underbrace{=}_\text{from Eq.~(\ref{eq:lgen})}\sum_{i=1}^n\mathbb{E}_{p(x_{1:n})}\left\{\log \frac{\sum_{y=1}^ce^{t_{\pi(i)}(y)/\tau}}{\int \sum_{y=1}^ce^{t_{\pi(i)}(y)/\tau}dx}\right\} \nonumber \\   &\qquad=\sum_{i=1}^n\mathbb{E}_{p(x_{\pi(i)})}\left\{\log \frac{\sum_{y=1}^ce^{t_{\pi(i)}(y)/\tau}}{\int \sum_{y=1}^ce^{t_{\pi(i)}(y)/\tau}dx}\right\} \nonumber\\       &\qquad\underbrace{=}_\text{$l\doteq\pi(i)$}\sum_{l=1}^n\mathbb{E}_{p(x_{l})}\left\{\log \frac{\sum_{y=1}^ce^{t_{l}(y)/\tau}}{\int \sum_{y=1}^ce^{t_{l}(y)/\tau}dx}\right\} \nonumber\\ 
    &\qquad =\mathcal{L}_{GEN}(\Theta,\mathcal{D}) \nonumber
\end{align}
Hence, failure mode 3 is an admissible solution.

Let's continue by proving statement b and recalling that 
\begin{align}
\mathcal{L}_{INV}(\Theta, \mathcal{D})=- \sum_{i=1}^n \mathbb{E}_{p(x_i)\mathcal{T}(x_i'|x_i)}\left\{CE(p(y_i|x_i';\Theta),p(y_i|x_i;\Theta))\right\} 
\label{eq:inv1}
\end{align}
Firstly, we test for representational collapse. In this case, we have that for all $i\in\{1,\dots,n\}$
\begin{align}  p(y_i|x_i;\Theta)=p(y_i|x_i';\Theta)=\text{Softmax}(f(k)/\tau) \nonumber
\end{align}
Based on this result, we observe that the cross-entropy terms in Eq.~(\ref{eq:inv1}) can be made 0 by proper choice of $k$. Therefore, representational collapse is an admissible solution. Secondly, we test for cluster collapse. Here, it is easy to see that the cross-entropy terms in Eq.~(\ref{eq:inv1}) are all 0. Therefore, also cluster collapse is admissible. Thirdly, we test for permutation invariance for the cluster assignments. On one hand, we have that the cross-entropy terms for $\mathcal{L}_{INV}(\Theta,\mathcal{D})$) in Eq.~(\ref{eq:inv1})  can be rewritten in the following way:
\begin{align}
    & CE(p(y_i|x_i';\Theta),p(y_i|x_i;\Theta))
    \nonumber\\
    &\qquad = CE\left(\frac{e^{t_i'(y_i)/\tau}}{\sum_{y=1}^ce^{t_i'(y)/\tau}},\frac{e^{t_i(y_i)/\tau}}{\sum_{y=1}^ce^{t_i(y)/\tau}}\right)
    \label{eq:ce1}
\end{align}
and the optimal solution is achieved only when $t_i'=t_i$ for all $i\in\{1,\dots,n\}$.
On the other hand, the cross-entropy terms for $\mathcal{L}_{INV}(\Theta,\mathcal{D}^\pi)$ are given by the following equality:
\begin{align}
    & CE(p(y_i|x_i';\Theta),p(y_i|x_i;\Theta))
    \nonumber\\
    &\qquad = CE\left(\frac{e^{t_i'(y_i)/\tau}}{\sum_{y=1}^ce^{t_i'(y)/\tau}},\frac{e^{t_{\pi(i)}(y_i)/\tau}}{\sum_{y=1}^ce^{t_{\pi(i)}(y)/\tau}}\right)
    \label{eq:ce2}
\end{align}
However, the optimal solution cannot be achieved in general as $t_i'\neq t_{\pi(i)}$ for some $i\in\{1,\dots,n\}$.\footnote{Indeed, note that $t_i'= t_{\pi(i)}$ for all $i$ occurs only when we are in one of the first two failure modes.} Therefore, $\mathcal{L}_{INV}$ is not permutation invariant to cluster assignments.

Let's conclude by proving statement c and recalling that
\begin{align}
    \mathcal{L}_{PRIOR}(\Theta,\mathcal{D}) = - \sum_{i=1}^nCE\left(p(y_i),\frac{1}{n}\sum_{l=1}^np(y_l=y_i|x_l;\Theta)\right)
    \label{eq:prior1}
\end{align}
Firstly, we test for representational collapse. One can easily observe that if $enc(x)=k$ for all $x\in\mathbb{R}^d$, $p(y|x;\Theta)$ becomes uniform, namely $p(y|x;\Theta)=1/c$ for all $y\in\{1,\dots,c\}$. Consequently, $\frac{1}{n}\sum_{l=1}^np(y_l=y_i|x_l;\Theta)=1/c$ for all $i\in\{1,\dots,n\}$. Now, since $p(y_i)=1/c$ for all $i\in\{1,\dots,n\}$, the cross-entropy terms in Eq.~(\ref{eq:prior1}) reach their maximum value $-H_p(y_i)$ for all $i\in\{1,\dots,n\}$. Therefore, representational collapse attains the global maximum of $\mathcal{L}_{PRIOR}$ and is an admissible solution. Secondly, we test for cluster collapse. By using the definition of cluster collapse, we observe that
\begin{align}
    \frac{1}{n}\sum_{l=1}^np(y_l=y_i|x_l;\Theta)=\left\{\begin{array}{lc}
        0 & y_i=j \\
        1 & y_i\neq j
    \end{array}\right.
\end{align}
Therefore, the resulting distribution is non-uniform, differently from $p(y_i)$. The cross-entropy terms in Eq.~(\ref{eq:prior1}) are not optimized and cluster collapse is not admissible. Thirdly, we test for permutation invariance of cluster assignments. We observe that
\begin{align}
    \frac{1}{n}\sum_{l=1}^np(y_l=y_i|x_l;\Theta)&=\frac{1}{n}\sum_{l=1}^n\frac{e^{t_l(y_i)/\tau}}{\sum_{y=1}^ce^{t_l(y)/\tau}} \nonumber\\
    &= \frac{1}{n}\sum_{l=1}^n\frac{e^{t_{\pi(l)}(y_i)/\tau}}{\sum_{y=1}^ce^{t_{\pi(l)}(y)/\tau}}
\end{align}
which is permutation invariant to cluster assignments. Consequently, also $\mathcal{L}_{PRIOR}(\Theta,\mathcal{D})=\mathcal{L}_{PRIOR}(\Theta,\mathcal{D}^\pi)$. This concludes the proof.
\end{proof}

{\section{Hyperparameters for Synthetic Data}\label{sec:hyperparams_synth}}
For the backbone $enc$, we use a MLP with two hidden layers and 100 neurons per layer, an output layer with 2 neurons and ReLU activation functions. For the projection head $proj$ ($f$ for GEDI and its variants), we use a MLP with one hidden layer and 4 neurons and an output layer with 2 neurons (batch normalization is used in all layers for Barlow and SwAV as required by their original formulation). All methods use a batch size of 400.
Baseline JEM (following the original paper):
\begin{itemize}
    \item Number of iterations $20K$
    \item Learning rate $1e-3$
    \item Optimizer Adam $\beta_1=0.9$, $\beta_2=0.999$
    \item SGLD steps $10$
    \item Buffer size $10000$
    \item Reinitialization frequency $0.05$
    \item SGLD step-size $\frac{0.01^2}{2}$
    \item SGLD noise $0.01$
\end{itemize}
And for self-supervised learning methods, please refer to Table~\ref{tab:hyperparams_synt_ssl}.

\begin{figure}
     \centering
     \begin{subfigure}[b]{0.45\linewidth}
         \centering
         \includegraphics[width=0.9\textwidth]{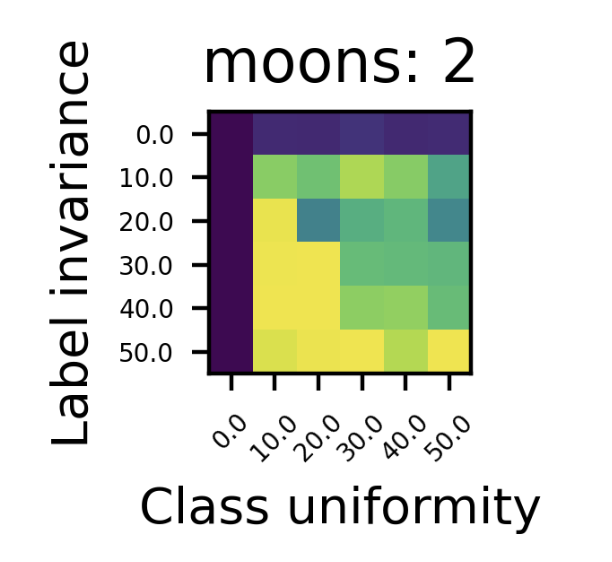}
         \caption{Moons}
     \end{subfigure}%
     \begin{subfigure}[b]{0.45\linewidth}
         \centering
         \includegraphics[width=0.9\textwidth]{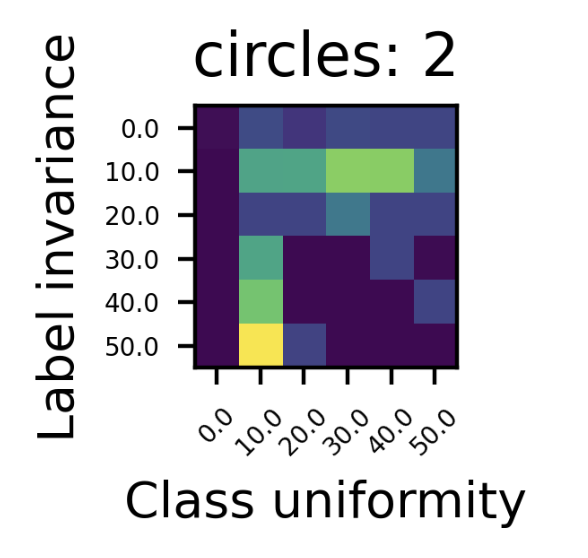}
         \caption{Circles}
     \end{subfigure}%
     \caption{Sensitivity analysis on the discriminative performance of GEDI for different loss weights (in the range $\{0,10,20,30,40,50\}$). Performance are averaged over 5 different random seeds. Yellow means perfect NMI.}
     \label{fig:sensitivity}
\end{figure}

We also provide an analysis of sensitivity to hyperparameters for GEDI. Please refer to Figure~\ref{fig:sensitivity}.
\begin{table*}
  \caption{Hyperparameters used in the synthetic experiments.}
  \label{tab:hyperparams_synt_ssl}
  \centering
  \begin{tabular}{lllll}
    \hline
    Methods & Barlow & SwAV & GEDI (no gen) & GEDI \\
    \hline
    Iters & \multicolumn{4}{c}{$20k$} \\
    Learning rate & \multicolumn{4}{c}{$1e-3$} \\
    Optimizer & \multicolumn{4}{c}{Adam $\beta_1=0.9, \beta_2=0.999$} \\
    Data augmentation noise $\sigma$ & \multicolumn{4}{c}{$0.03$} \\
    SGLD steps $T$ & - & - & 1 & 1 \\
    Buffer size $|B|$ & - & - & 10000 & 10000 \\
    Reinitialization frequenc & - & - & 0.05 & 0.05 \\
    SGLD step size & - & - & $\frac{0.01^2}{2}$ $\frac{0.01^2}{2}$ & $\frac{0.01^2}{2}$ \\        
    SGLD noise & - & - & 0.01 & 0.01 \\
    Weight for $\mathcal{L}_{GEN}$ & - & - & $1$ & $1$ \\
    Weight for $\mathcal{L}_{INV}$ & - & - & $50$ & $50$ \\
    Weight for $\mathcal{L}_{PRIOR}$ & - & - & $10$ & $10$ \\
    \hline
  \end{tabular}
\end{table*}

\begin{figure*}
     \centering
     \begin{subfigure}[b]{0.14\linewidth}
         \centering
         \includegraphics[width=0.9\textwidth]{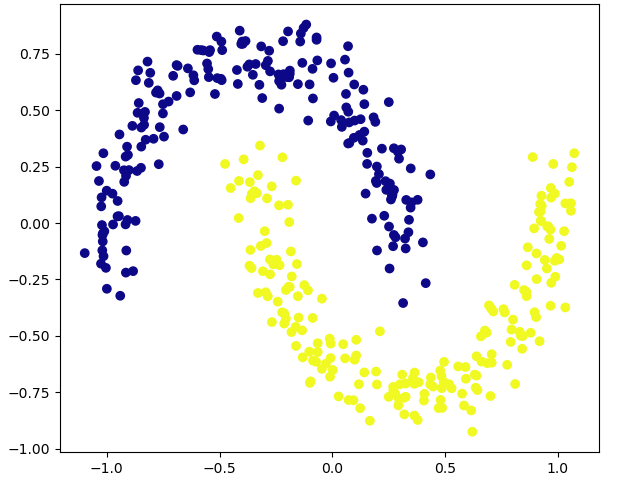}
         \caption{Truth}
     \end{subfigure}%
     \begin{subfigure}[b]{0.14\linewidth}
         \centering
         \includegraphics[width=0.9\textwidth]{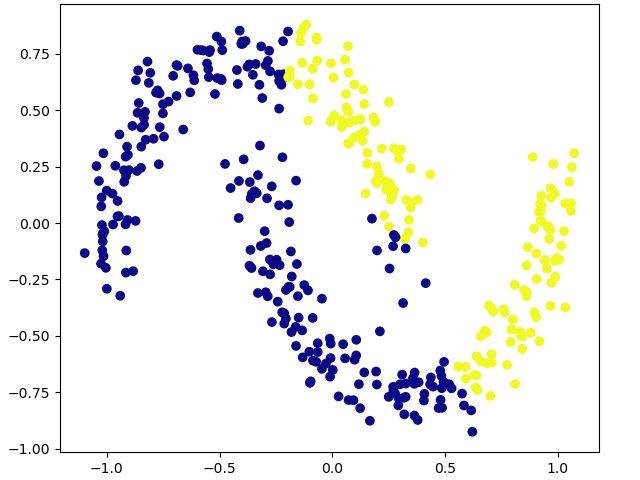}
         \caption{Barlow}
     \end{subfigure}%
     \begin{subfigure}[b]{0.14\linewidth}
         \centering
         \includegraphics[width=0.9\textwidth]{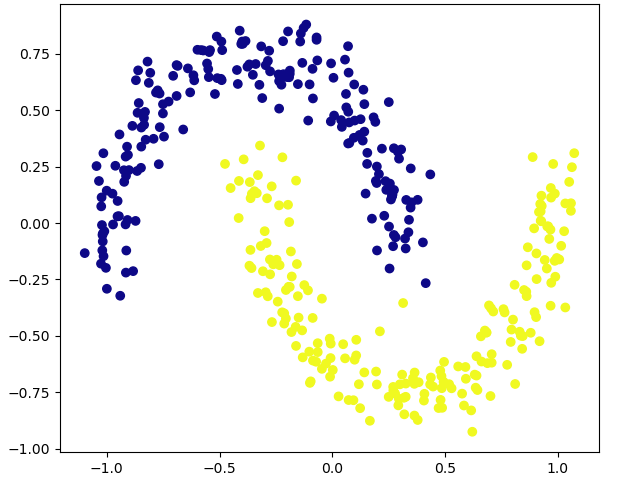}
         \caption{SwAV}
     \end{subfigure}%
     \begin{subfigure}[b]{0.14\linewidth}
         \centering
         \includegraphics[width=0.9\textwidth]{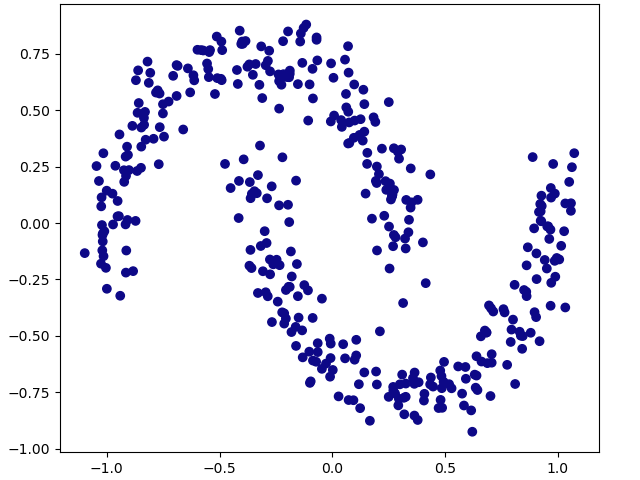}
         \caption{no unif}
     \end{subfigure}%
     \begin{subfigure}[b]{0.14\linewidth}
         \centering
         \includegraphics[width=0.9\textwidth]{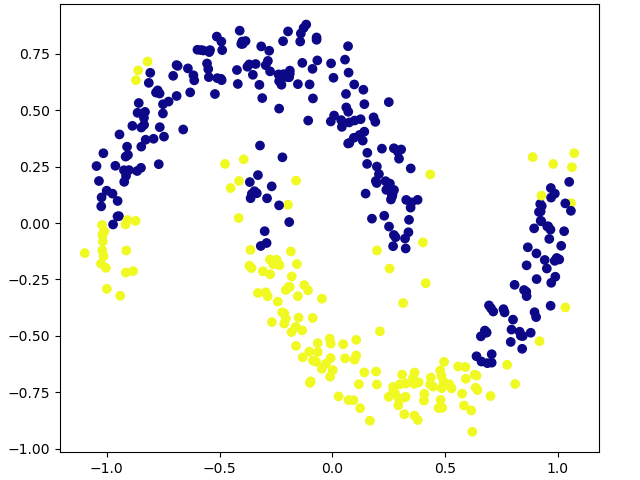}
         \caption{no inv}
     \end{subfigure}%
     \begin{subfigure}[b]{0.14\linewidth}
         \centering
         \includegraphics[width=0.9\textwidth]{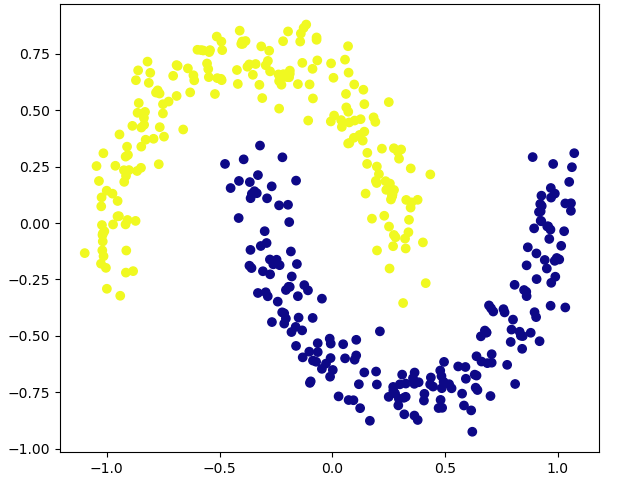}
         \caption{no gen}
     \end{subfigure}%
     \begin{subfigure}[b]{0.14\linewidth}
         \centering
         \includegraphics[width=0.9\textwidth]{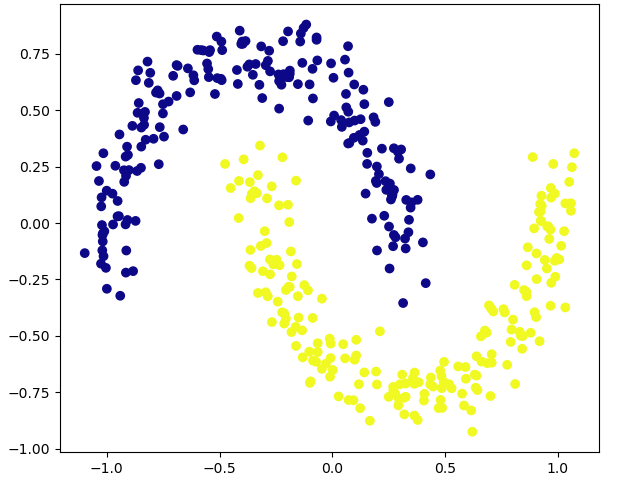}
         \caption{GEDI}
     \end{subfigure}%
     \\
     \begin{subfigure}[b]{0.14\linewidth}
         \centering
         \includegraphics[width=0.9\textwidth]{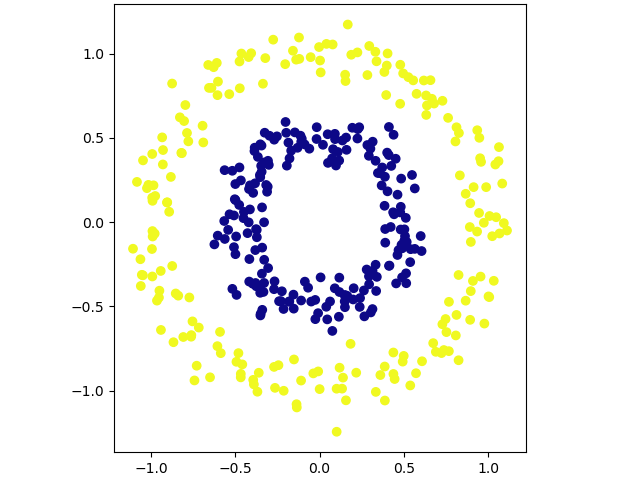}
         \caption{Truth}
     \end{subfigure}%
     \begin{subfigure}[b]{0.14\linewidth}
         \centering
         \includegraphics[width=0.9\textwidth]{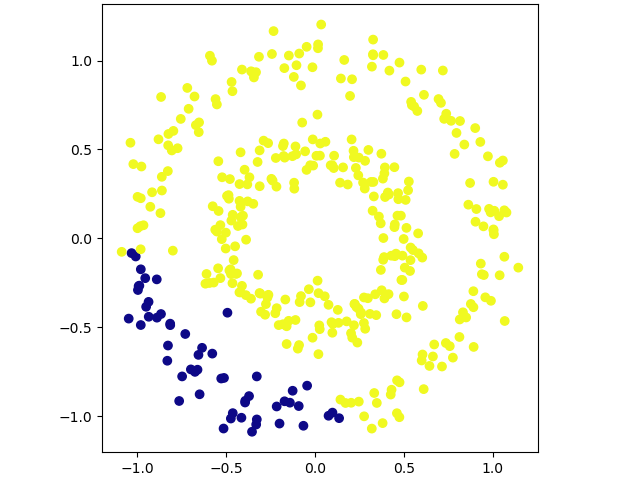}
         \caption{Barlow}
     \end{subfigure}%
     \begin{subfigure}[b]{0.14\linewidth}
         \centering
         \includegraphics[width=0.9\textwidth]{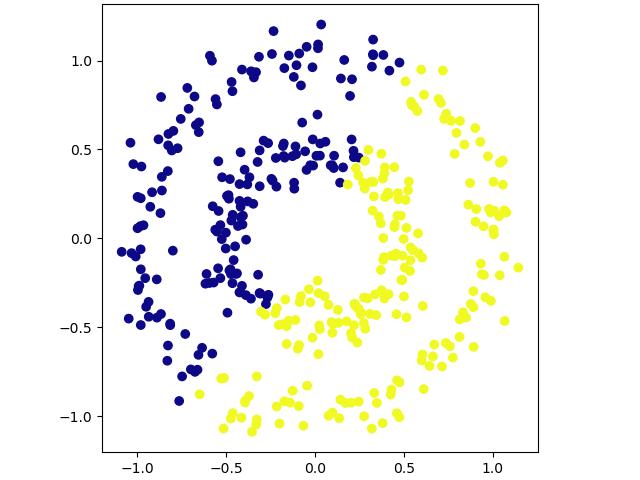}
         \caption{SwAV}
     \end{subfigure}%
     \begin{subfigure}[b]{0.14\linewidth}
         \centering
         \includegraphics[width=0.9\textwidth]{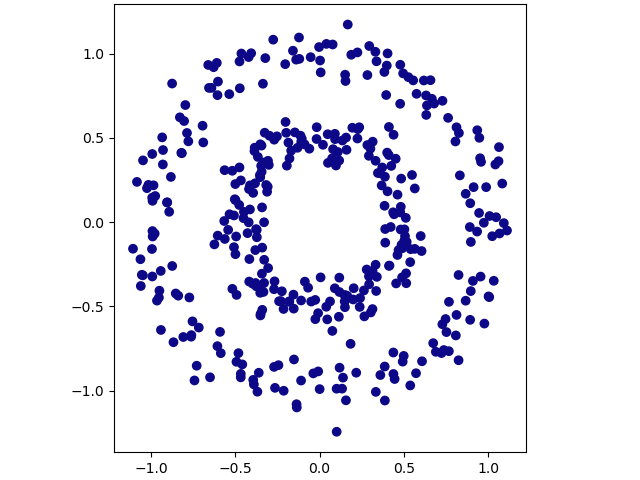}
         \caption{no unif}
     \end{subfigure}%
     \begin{subfigure}[b]{0.14\linewidth}
         \centering
         \includegraphics[width=0.9\textwidth]{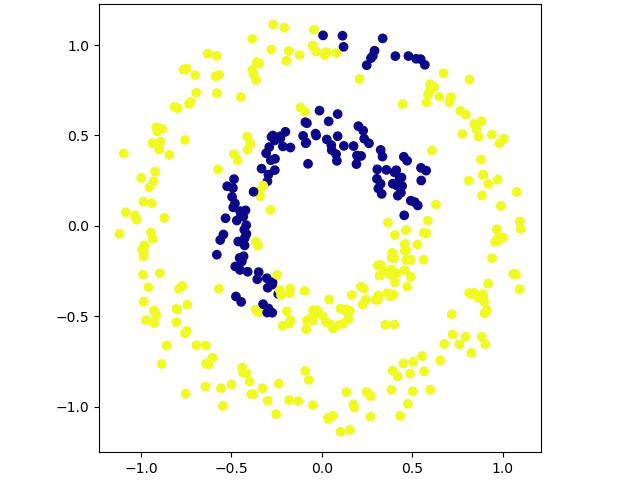}
         \caption{no inv}
     \end{subfigure}%
     \begin{subfigure}[b]{0.15\linewidth}
         \centering
         \includegraphics[width=0.9\textwidth]{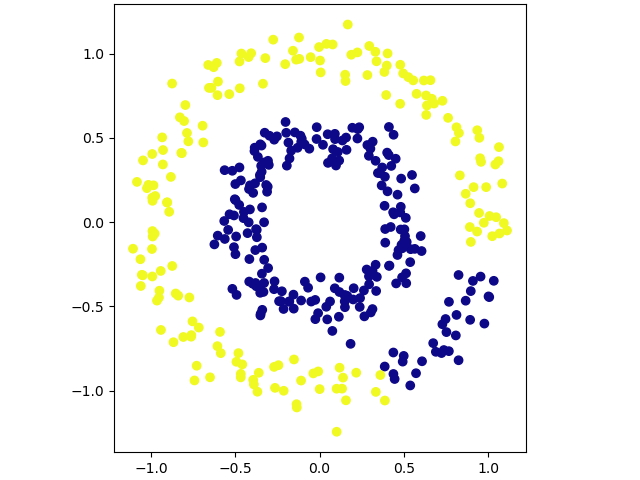}
         \caption{no gen}
     \end{subfigure}%
     \begin{subfigure}[b]{0.14\linewidth}
         \centering
         \includegraphics[width=0.9\textwidth]{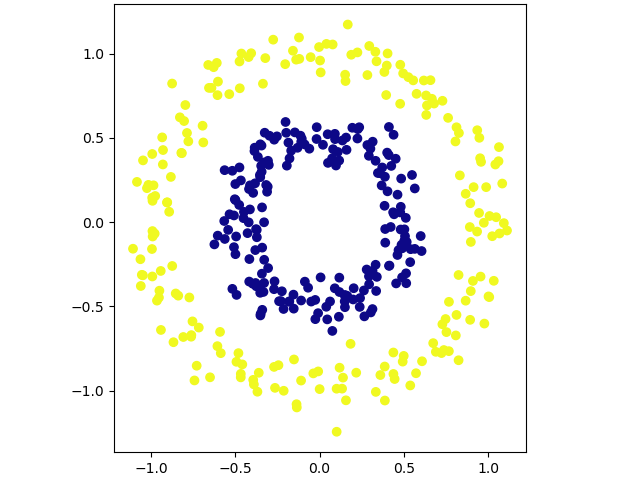}
         \caption{GEDI}
     \end{subfigure}%
     \caption{Qualitative visualization of the clustering performance for the different strategies on moons (a-g) and on circles (h-n) datasets. Colors identify different cluster predictions. Only GEDI and GEDI (no gen) are able to perform well on both datasets.}
     \label{fig:labels_toy}
\end{figure*}

\begin{figure*}
     \centering
     \begin{subfigure}[b]{0.14\linewidth}
         \centering
         \includegraphics[width=0.9\textwidth]{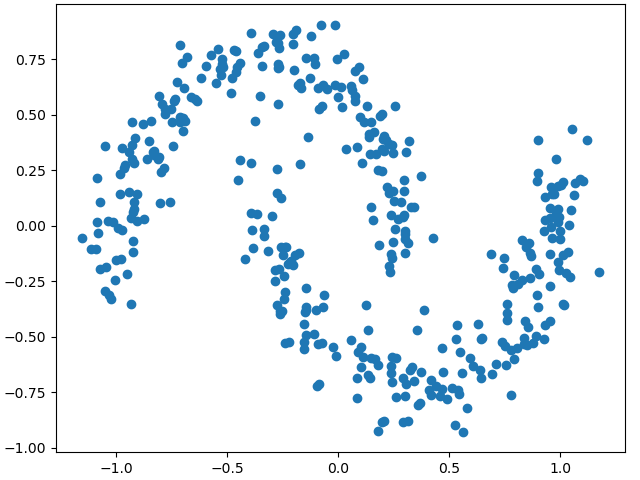}
         \caption{JEM}
     \end{subfigure}%
     \begin{subfigure}[b]{0.14\linewidth}
         \centering
         \includegraphics[width=0.9\textwidth]{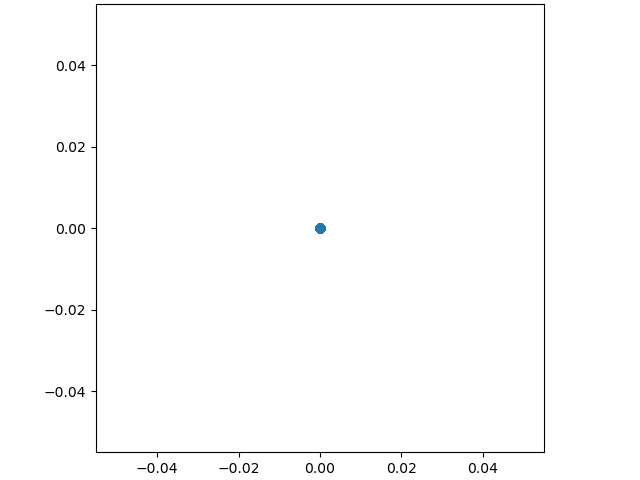}
         \caption{Barlow}
     \end{subfigure}%
     \begin{subfigure}[b]{0.14\linewidth}
         \centering
         \includegraphics[width=0.9\textwidth]{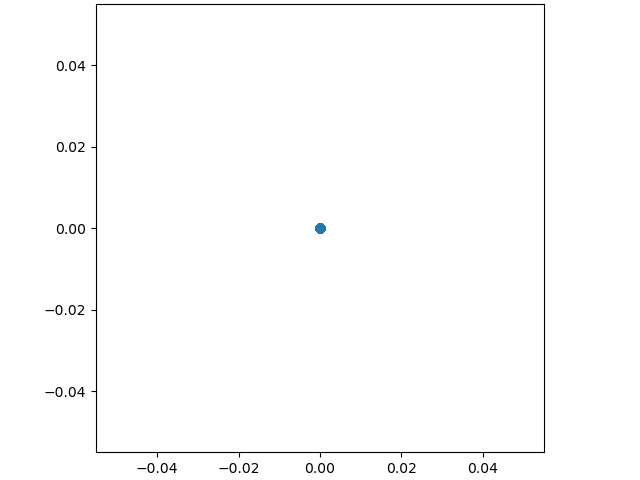}
         \caption{SwAV}
     \end{subfigure}%
     \begin{subfigure}[b]{0.14\linewidth}
         \centering
         \includegraphics[width=0.9\textwidth]{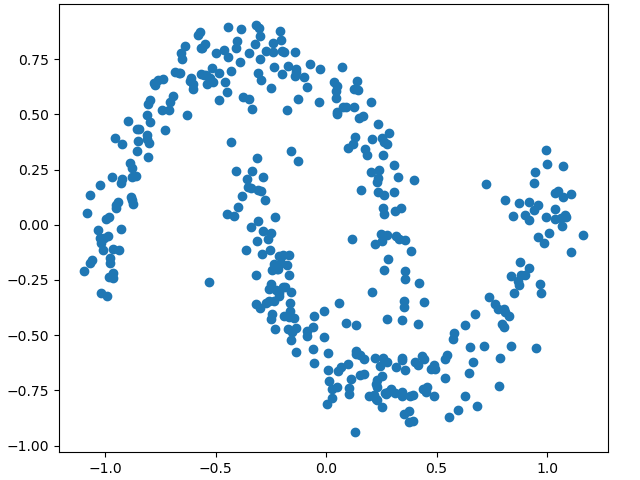}
         \caption{no unif}
     \end{subfigure}%
     \begin{subfigure}[b]{0.14\linewidth}
         \centering
         \includegraphics[width=0.9\textwidth]{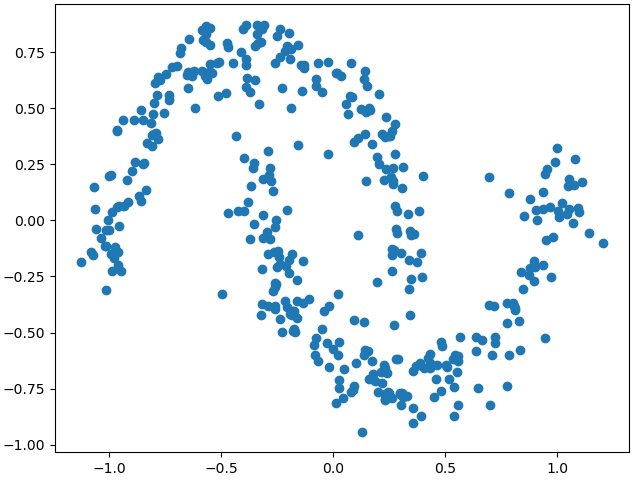}
         \caption{no inv}
     \end{subfigure}%
     \begin{subfigure}[b]{0.14\linewidth}
         \centering
         \includegraphics[width=0.9\textwidth]{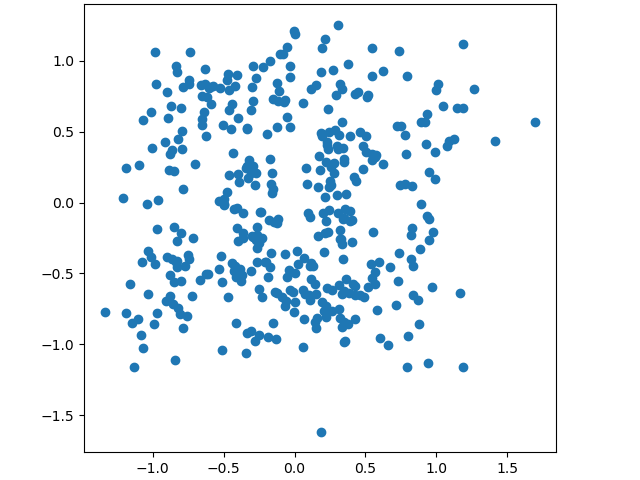}
         \caption{no gen}
     \end{subfigure}%
     \begin{subfigure}[b]{0.14\linewidth}
         \centering
         \includegraphics[width=0.9\textwidth]{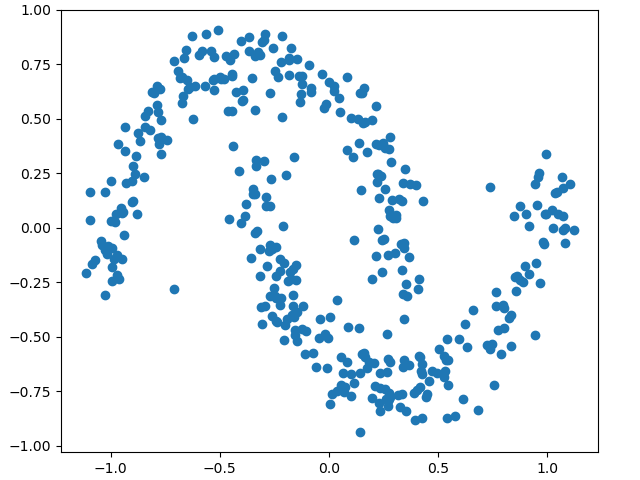}
         \caption{GEDI}
     \end{subfigure}%
     \\
     \begin{subfigure}[b]{0.14\linewidth}
         \centering
         \includegraphics[width=0.9\textwidth]{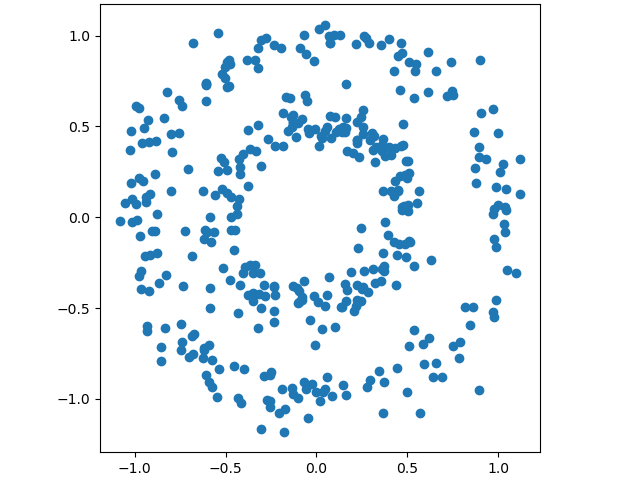}
         \caption{JEM}
     \end{subfigure}%
     \begin{subfigure}[b]{0.14\linewidth}
         \centering
         \includegraphics[width=0.9\textwidth]{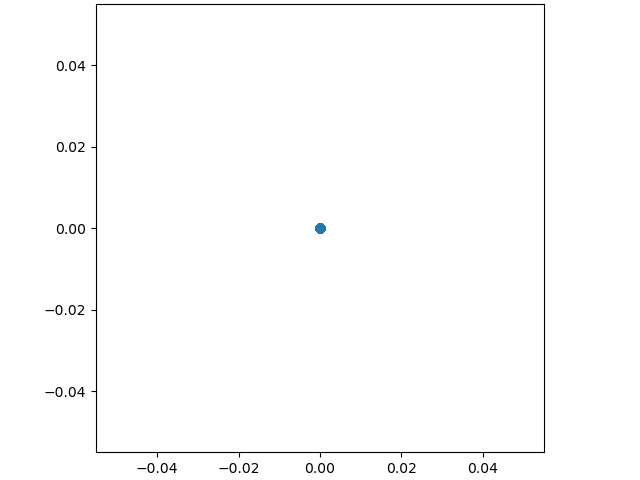}
         \caption{Barlow}
     \end{subfigure}%
     \begin{subfigure}[b]{0.14\linewidth}
         \centering
         \includegraphics[width=0.9\textwidth]{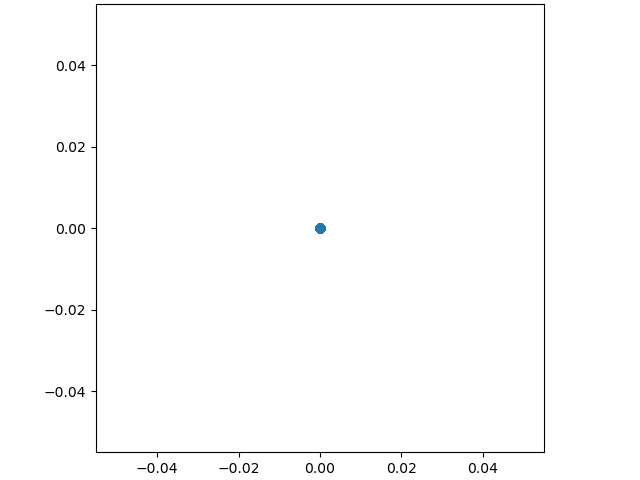}
         \caption{SwAV}
     \end{subfigure}%
     \begin{subfigure}[b]{0.14\linewidth}
         \centering
         \includegraphics[width=0.9\textwidth]{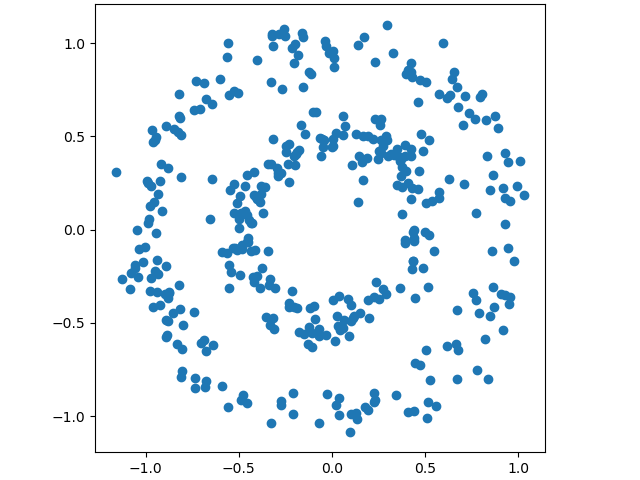}
         \caption{no unif}
     \end{subfigure}%
     \begin{subfigure}[b]{0.14\linewidth}
         \centering
         \includegraphics[width=0.9\textwidth]{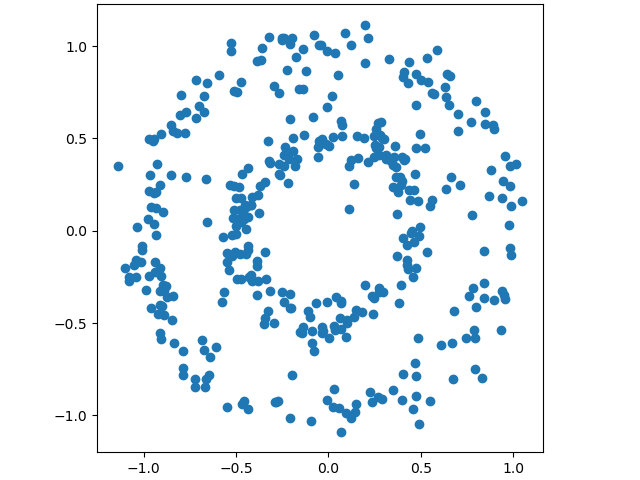}
         \caption{no inv}
     \end{subfigure}%
     \begin{subfigure}[b]{0.15\linewidth}
         \centering
         \includegraphics[width=0.9\textwidth]{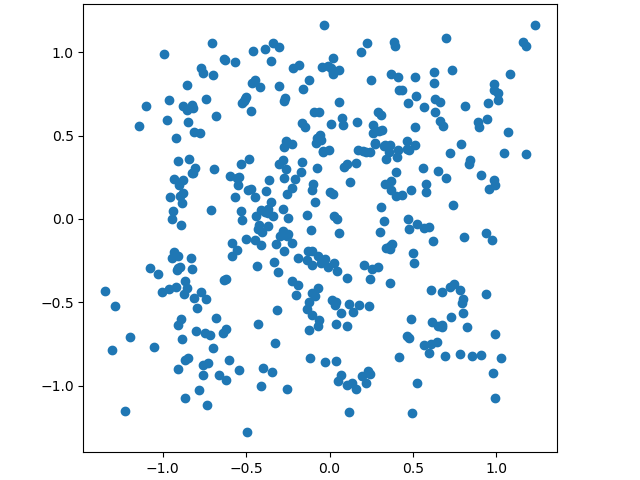}
         \caption{no gen}
     \end{subfigure}%
     \begin{subfigure}[b]{0.14\linewidth}
         \centering
         \includegraphics[width=0.9\textwidth]{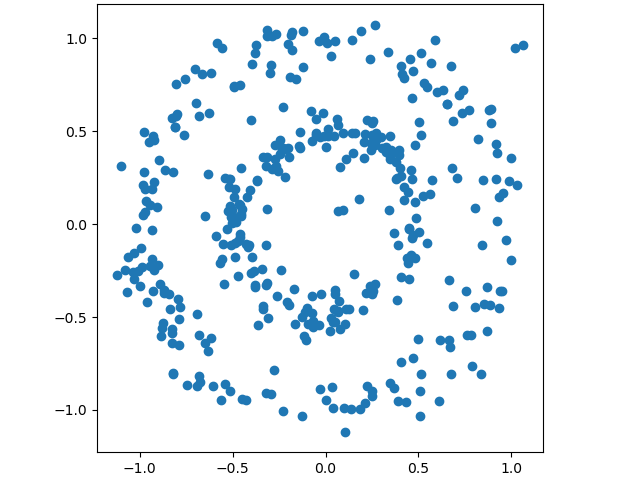}
         \caption{GEDI}
     \end{subfigure}%
     \caption{Qualitative visualization of the generative performance for the different strategies on moons (a-g) and on circles (h-n) datasets. Colors identify different cluster predictions. All GEDI approaches (except for no gen perform comparably well to the generative solution JEM).}
     \label{fig:gen_toy}
\end{figure*}

{\section{Additional Experiments for Toy Data}}
Please, refer to Figure~\ref{fig:labels_toy} for the discriminative performance and Figure~\ref{fig:gen_toy} for the generative ones.

{\section{Hyperparameters for SVHN, CIFAR10, CIFAR100}\label{sec:hyperparams_real}}
\begin{table}[ht]
  \caption{Resnet architecture. Conv2D(A,B,C) applies a 2d convolution to input with B channels and produces an output with C channels using stride (1, 1), padding (1, 1) and kernel size (A, A).}
  \label{tab:backbone}
  \centering
\begin{tabular}{@{}lrr@{}}
\toprule
\textbf{Name} & \textbf{Layer} & \textbf{Res. Layer} \\
\midrule
\multirow{6}{*}{Block 1} & Conv2D(3,3,F) & \multirow{2}{*}{AvgPool2D(2)} \\
& LeakyRELU(0.2) & \\
& Conv2D(3,F,F) & \multirow{2}{*}{Conv2D(1,3,F) no padding}\\
& AvgPool2D(2) & \\
& \cline{1-2}\\
& \multicolumn{2}{c}{Sum} \\
\midrule
\multirow{5}{*}{Block 2} & LeakyRELU(0.2) & \\
& Conv2D(3,F,F) & \\
& LeakyRELU(0.2) & \\
& Conv2D(3,F,F) & \\
& AvgPool2D(2) & \\
\midrule
\midrule
\multirow{4}{*}{Block 3} & LeakyRELU(0.2) & \\
& Conv2D(3,F,F) & \\
& LeakyRELU(0.2) & \\
& Conv2D(3,F,F) & \\
\midrule
\multirow{5}{*}{Block 4} & LeakyRELU(0.2) & \\
& Conv2D(3,F,F) & \\
& LeakyRELU(0.2) & \\
& Conv2D(3,F,F) & \\
& AvgPool2D(all) & \\
\bottomrule
\end{tabular}
\end{table}
For the backbone $enc$, we use a ResNet with 8 layers as in~\cite{duvenaud2021no}, where its architecture is shown in Table~\ref{tab:backbone}. For the projection head $proj$ ($f$ for GEDI and its variants), we use a MLP with one hidden layer and $2*F$ neurons and an output layer with $F$ neurons (batch normalization is used in all layers for Barlow and SwAV as required by their original formulation + final $L_2$ normalization). $F=128$ for SVHN, CIFAR-10 (1 million parameters) and $F=256$ for CIFAR-100 (4.1 million parameters). For JEM, we use the same settings of~\cite{duvenaud2021no}. All methods use a batch size of 64.
Baseline JEM (following the original paper):
\begin{itemize}
    \item Number of epochs $20$, $200$, $200$ for SVHN, CIFAR-10, CIFAR-100, respectively.
    \item Learning rate $1e-4$
    \item Optimizer Adam
    \item SGLD steps $20$
    \item Buffer size $10000$
    \item Reinitialization frequency $0.05$
    \item SGLD step-size $1$
    \item SGLD noise $0.01$
    \item Data augmentation (Gaussian noise) $0.03$
\end{itemize}
And for self-supervised learning methods, please refer to Table~\ref{tab:hyperparams}.

\begin{table*}[ht]

  \caption{Hyperparameters (in terms of sampling, optimizer, objective and data augmentation) used in all experiments.}
  \label{tab:hyperparams}
  \centering
\resizebox{\linewidth}{!}{\begin{tabular}{@{}llrrr@{}}
\toprule
\textbf{Class} & \textbf{Name param.} & \textbf{SVHN} & \textbf{CIFAR-10} & \textbf{CIFAR-100} \\
\midrule
\multirow{5}{*}{Data augment.} & Color jitter prob. & 0.1 & 0.1 & 0.1 \\
& Gray scale prob. & 0.1 & 0.1 & 0.1 \\
& Random crop & Yes & Yes & Yes \\
& Additive Gauss. noise (std) & 0.03 & 0.03 & 0.03 \\
& Random horizontal flip & No & Yes & Yes \\
\midrule
\multirow{5}{*}{SGLD} & SGLD iters & 20 & 20 & 20 \\
& Buffer size & 10k & 10k & 10k \\
& Reinit. frequency & 0.05 & 0.05 &0.05 \\
& SGLD step-size & 1 & 1 & 1 \\
& SGLD noise & 0.01 & 0.01 & 0.01 \\
\midrule
\multirow{6}{*}{Optimizer} & Batch size & 64 & 64 & 64  \\
& Epochs & 20 & 200 & 200 \\
& Adam $\beta_1$ & 0.9 & 0.9 & 0.9 \\
& Adam $\beta_2$ & 0.999 & 0.999 & 0.999  \\
& Learning rate & $1e-4$ & $1e-4$ & $1e-4$ \\
\midrule
\multirow{5}{*}{Weights for losses} & $\mathcal{L}_{GEN}$ & 1 & 1 & 1 \\
& $\mathcal{L}_{INV}$ & 50 & 50 & 50 \\
& $\mathcal{L}_{PRIOR}$ & 25 & 25 & 50 \\
& $\mathcal{L}_{NeSY}$ & - & - & - \\
\bottomrule
\end{tabular}}
\end{table*}

{\section{Experiments on SVHN, CIFAR-10, CIFAR-100}\label{sec:additional}}
\begin{table}[t]
  \caption{Generative and discriminative performance on test set (SVHN, CIFAR-10, CIFAR-100). Normalized mutual information  (NMI) and Frechet Inception Distance (FID) are used as evaluation metrics for the discriminative and generative tasks, respectively. Higher values of NMI and lower values of FID indicate better performance.}
  \label{tab:nmi_real}
  \centering
\begin{tabular}{@{}llrrr@{}}
\toprule
\textbf{Task} & \textbf{Method} & \textbf{SVHN} & \textbf{CIFAR-10} & \textbf{CIFAR-100}   \\
\midrule
\multirow{5}{*}{Discriminative} & JEM & 0.00 & 0.00 & 0.00 \\
& Barlow & 0.23 & 0.17 & 0.58 \\
& SwAV & 0.21 & 0.43 & 0.65 \\
(NMI) & GEDI (no gen) & 0.21 & 0.44 & 0.86 \\
& GEDI & \textbf{0.25} & \textbf{0.45} & \textbf{0.87} \\
\midrule
\multirow{5}{*}{Generative} & JEM & \textbf{201} & 263 & 226 \\
& Barlow & 353 & 392 & 352 \\
& SwAV & 584 & 415 & 416 \\
(FID) & GEDI (no gen) & 454 & 424 & 447 \\
& GEDI & 218 & \textbf{197} & \textbf{222} \\
\bottomrule
\end{tabular}
\end{table}
\begin{table}
  \caption{OOD detection in terms of AUROC on test set (CIFAR-10, CIFAR-100). Training is performed on SVHN.}
  \label{tab:oodsvhn}
  \centering
  \begin{tabular}{@{}llllll@{}}
        \toprule
        \textbf{Dataset} & \textbf{JEM} & \textbf{Barlow} & \textbf{SwAV} & \textbf{GEDI no gen} & \textbf{GEDI} \\
    \midrule
    CIFAR-10 & 0.73 & 0.17 & 0.26 & 01 & \textbf{0.80}  \\
    CIFAR-100 & 0.72 & 0.24 & 0.32 & 0.15 & \textbf{0.80} \\
    \bottomrule
  \end{tabular}
\end{table}
\begin{figure}
     \centering
     \begin{subfigure}[b]{0.45\linewidth}
         \centering
         \includegraphics[width=0.9\textwidth]{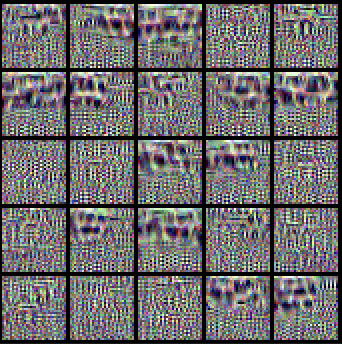}
         \caption{Barlow}
     \end{subfigure}%
     \begin{subfigure}[b]{0.45\linewidth}
         \centering
         \includegraphics[width=0.9\textwidth]{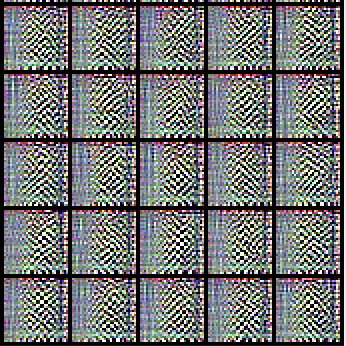}
         \caption{SwAV}
     \end{subfigure}%
     \\
     \begin{subfigure}[b]{0.45\linewidth}
         \centering
         \includegraphics[width=0.9\textwidth]{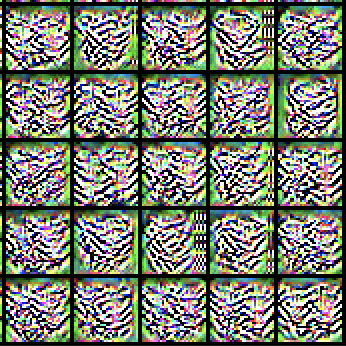}
         \caption{GEDI no gen}
     \end{subfigure}%
     \begin{subfigure}[b]{0.45\linewidth}
         \centering
         \includegraphics[width=0.9\textwidth]{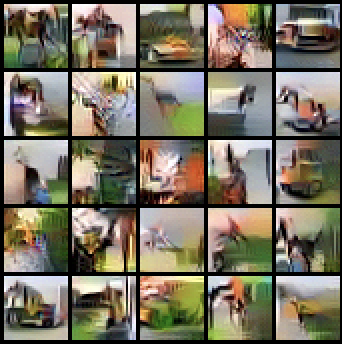}
         \caption{GEDI}
     \end{subfigure}%
     \caption{Qualitative visualization of the generative performance for the different discriminative strategies on CIFAR-10. Results are obtained by running Stochastic Langevin Dynamics for 500 iterations.}
     \label{fig:samples}
\end{figure}
We consider three well-known computer vision benchmarks, namely SVHN, CIFAR-10 and CIFAR-100. We use a simple 8-layer Resnet network for the backbone encoder for both SVHN and CIFAR-10 (around 1M parameters) and increase the hidden layer size for CIFAR-100 (around 4.1M parameters) as from~\cite{duvenaud2021no}. We use a MLP with a single hidden layer for $proj$ (the number of hidden neurons is double the number of inputs), we choose $h=256$ for CIFAR-100 and $h=128$ for all other cases. Additionally, we use data augmentation strategies commonly used in the SSL literature, including color jitter, and gray scale conversion to name a few. We train JEM, Barlow, SwAV, GEDI no gen and GEDI using Adam optimizer with learning rate $1e-4$ and batch size $64$ for $20$, $200$ and $200$ epochs for each respective dataset (SVHN, CIFAR-10 AND CIFAR-100). Further details about the hyperparameters are available in the Supplementary Material (Section I). Similarly to the toy experiments, we evaluate the clustering performance by using the Normalized Mutual Information (NMI) score. Additionally, we evaluate the generative performance qualitatively using the Frechet Inception Distance~\cite{heusel2017gans} as well as the OOD detection capabilities following the methodology in~\cite{grathwohl2020your}.

From Table~\ref{tab:nmi_real}, we observe that GEDI is able to outperform all other competitors by a large margin, thanks to the properties of both generative and self-supervised models. We observe that the difference gap in clustering performance increases with a larger number of classes (cf. CIFAR-100). This might be explained by the fact that the number of possible label permutations can increase with the number of classes and that our loss is more robust to the permutation invariance problem as from Theorem~\ref{thm:admissible}. We observe also that GEDI no gen is comparable and often superior to SwAV, despite being simpler (i.e. avoiding the use of asymmetries and the running of iterative clustering). In terms of generation performance, GEDI is the only approach that compares favorably with JEM. We provide a qualitative set of samples generated by the different discriminative models in Figure~\ref{fig:samples}.

Last but not least, we investigate the OOD detection capabilities of the different methods. Table~\ref{tab:oodCIFAR-10} provides a quantitative summary of the performance for a subset of experiments (the complete set is available in Section J). We observe that GEDI is more robust compared to other discriminative baselines, thanks to its generative nature.

Overall, these experiments provide real-world evidence on the benefits of the proposed unification and theoretical results.

We conduct a linear probe evaluation of the representations learnt by the different models Table~\ref{tab:acc}. These experiments provide insights on the capabilities of learning representations producing linearly separable classes. From Table~\ref{tab:acc}, we observe a large difference in results between Barlow and SwAV. Our approach provides interpolating results between the two approaches.

\begin{figure}
     \centering
     \begin{subfigure}[b]{0.45\linewidth}
         \centering
         \includegraphics[width=0.9\textwidth]{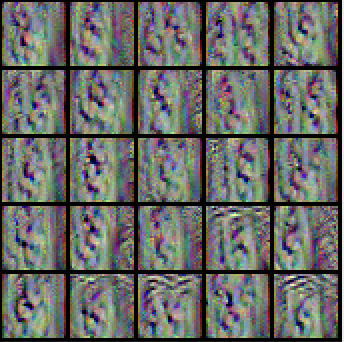}
         \caption{Barlow}
     \end{subfigure}%
     \begin{subfigure}[b]{0.45\linewidth}
         \centering
         \includegraphics[width=0.9\textwidth]{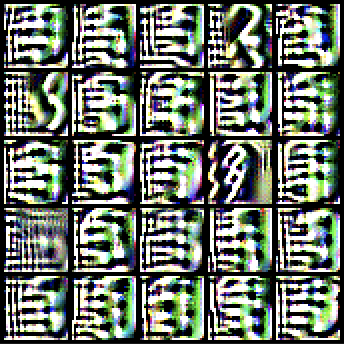}
         \caption{SwAV}
     \end{subfigure}%
     \\
     \begin{subfigure}[b]{0.45\linewidth}
         \centering
         \includegraphics[width=0.9\textwidth]{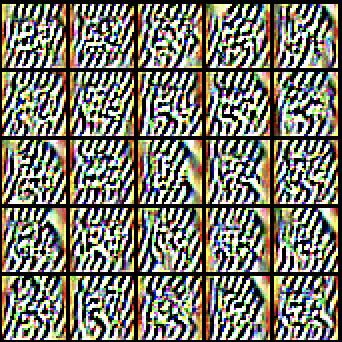}
         \caption{GEDI no gen}
     \end{subfigure}%
     \begin{subfigure}[b]{0.45\linewidth}
         \centering
         \includegraphics[width=0.9\textwidth]{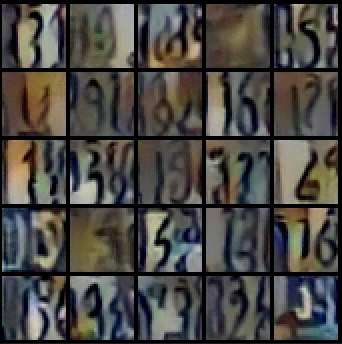}
         \caption{GEDI}
     \end{subfigure}%
     \caption{Qualitative visualization of the generative performance for the different discriminative strategies on SVHN. Results are obtained by running Stochastic Langevin Dynamics for 500 iterations.}
     \label{fig:samples_2}
\end{figure}

\begin{figure}
     \centering
     \begin{subfigure}[b]{0.45\linewidth}
         \centering
         \includegraphics[width=0.9\textwidth]{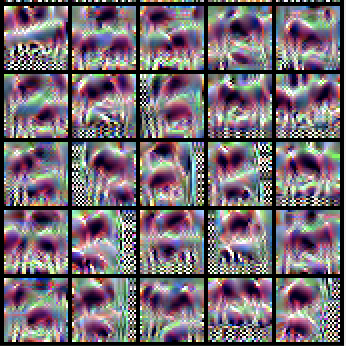}
         \caption{Barlow}
     \end{subfigure}%
     \begin{subfigure}[b]{0.45\linewidth}
         \centering
         \includegraphics[width=0.9\textwidth]{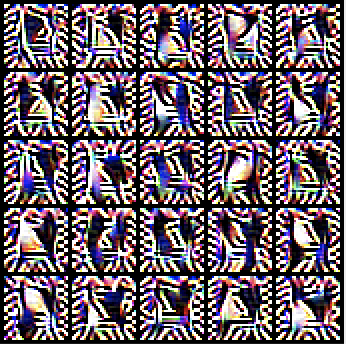}
         \caption{SwAV}
     \end{subfigure}%
     \\
     \begin{subfigure}[b]{0.45\linewidth}
         \centering
         \includegraphics[width=0.9\textwidth]{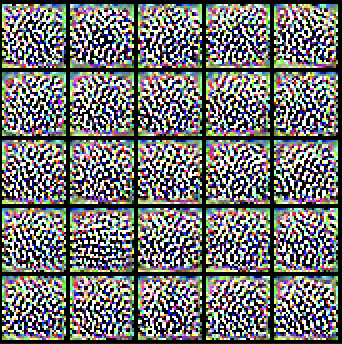}
         \caption{GEDI no gen}
     \end{subfigure}%
     \begin{subfigure}[b]{0.45\linewidth}
         \centering
         \includegraphics[width=0.9\textwidth]{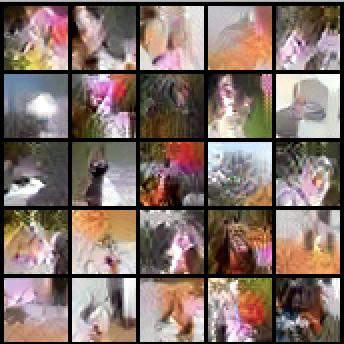}
         \caption{GEDI}
     \end{subfigure}%
     \caption{Qualitative visualization of the generative performance for the different discriminative strategies on CIFAR-100. Results are obtained by running Stochastic Langevin Dynamics for 500 iterations.}
     \label{fig:samples_3}
\end{figure}

We also provide additional qualitative analyisis on the generation performance on SVHN and CIFAR-100. Please, refer to Figure~\ref{fig:samples_2} and Figure~\ref{fig:samples_3}.

Finally, we evaluate the performance in terms of OOD detection, by following the same methodology used in~\cite{grathwohl2020your}. We use the OOD score criterion proposed in~\cite{grathwohl2020your}, namely $s(x)=-\|\frac{\partial\log p_\Psi(x)}{\partial x}\|_2$. From Table~\ref{tab:oodsvhn}, we observe that GEDI achieves almost optimal performance. While these results are exciting, it is important to mention that they are not generally valid. Indeed, when training on CIFAR-10 and performing OOD evaluation on the other datasets, we observe that all approaches achieve similar performance both on CIFAR-100 and SVHN, suggesting that all datasets are considered in-distribution, see Table~\ref{tab:oodCIFAR-10}. A similar observation is obtained when training on CIFAR-100 and evaluating on CIFAR-10 and SVHN, see Table~\ref{tab:oodCIFAR-100}. Importantly, this is a phenomenon which has been only recently observed by the scientific community on generative models. Tackling this problem is currently out of the scope of this work. For further discussion about the issue, we point the reader to the works in~\cite{nalisnick2019deep}.
\begin{table}[h]
  \caption{Supervised linear evaluation in terms of accuracy on test set (SVHN, CIFAR-10, CIFAR-100). The linear classifier is trained for 100 epochs using SGD with momentum, learning rate $1e-3$ and batch size 100.}
  \label{tab:acc}
  \centering
  \begin{tabular}{@{}llllll@{}}
    \toprule
    \textbf{Dataset} & \textbf{JEM} & \textbf{Barlow} & \textbf{SwAV} & \textbf{GEDI no gen} & \textbf{GEDI} \\
    \midrule
    SVHN & 0.20 & 0.74 & 0.45 & 0.35  & 0.55 \\
    CIFAR-10 & 0.23 & 0.65 & 0.46 & 0.54 & 0.53 \\
    CIFAR-100 & 0.03 & 0.27 & 0.13  & 0.14 & 0.12 \\
    \bottomrule
  \end{tabular}
\end{table}
\begin{table}
  \caption{OOD detection in terms of AUROC on test set (SVHN, CIFAR-100). Training is performed on CIFAR-10.}
  \label{tab:oodCIFAR-10}
  \centering
  \begin{tabular}{@{}llllll@{}}
        \toprule
        \textbf{Dataset} & \textbf{JEM} & \textbf{Barlow} & \textbf{SwAV} & \textbf{GEDI no gen} & \textbf{GEDI} \\
    \midrule
    SVHN & 0.44 & 0.32 & \textbf{0.62} & 0.11 & 0.57  \\
    CIFAR-100 & 0.53 & 0.56 & 0.51 & 0.51 & \textbf{0.61}  \\
    \bottomrule
  \end{tabular}
\end{table}

\begin{table}
  \caption{OOD detection in terms of AUROC on test set (SVHN, CIFAR-10). Training is performed on CIFAR-100.}
  \label{tab:oodCIFAR-100}
  \centering
  \begin{tabular}{@{}llllll@{}}
        \toprule
        \textbf{Dataset} & \textbf{JEM} & \textbf{Barlow} & \textbf{SwAV} & \textbf{GEDI no gen} & \textbf{GEDI} \\
    \midrule
    SVHN & 0.44 & 0.45 & 0.3 & \textbf{0.55} & 0.53 \\
    CIFAR-10 & \textbf{0.49} & 0.43 & 0.47 & 0.46 & 0.48 \\
    \bottomrule
  \end{tabular}
\end{table}

\end{document}